# Analyzing the Impact of Sentiments of Scientific Articles on COVID-19 Vaccination Rates

Sean Eugene G. Chua (seaneugene.chua@mail.utoronto.ca)
Kevin Anthony S. Sison - St. Luke's Medical Center - College of Medicine WHQM
(sison.ka.f@slmc-cm.edu.ph)

**Abstract**

At the peak of the COVID-19 pandemic, numerous countries worldwide sought to mobilize vaccination campaigns in an attempt to curb the spread and number of deaths caused by the virus. One avenue in which information regarding COVID vaccinations is propagated is that of scientific articles, which provide a certain level of credibility regarding this. Hence, this increases the probability that people who view these articles would get vaccinated if the articles convey a positive message on vaccinations and conversely decreases the probability of vaccinations if the articles convey a negative message. This being said, this study aims to investigate the correlation between article sentiments and the corresponding increase or decrease in vaccinations in the United States. To do this, a lexicon-based sentiment analysis was performed in two steps: first, article content was scraped via a Python library called BeautifulSoup, and second, VADER was used to obtain the sentiment analysis scores for each article based on the scraped text content. Results suggest that there was a relatively weak correlation between the average sentiment score of articles and the corresponding increase or decrease in COVID vaccination rates in the US.

**Introduction**

The SARS-Cov2 pandemic, commonly known as COVID-19 and officially declared as a global pandemic by the World Health Organization (WHO) on March 11, 2020, has brought about advances within the medical field, especially in pertinence to vaccine production. Due to the urgency of the COVID-19 situation around the world, scientists, researchers, and pharmaceutical companies sought to fast-track the production of the different COVID vaccines we know today. These vaccines vary in form from mRNA (Pfizer-BioNTech and Moderna), viral vector (Johnson & Johnson Janssen), and live attenuated virus (CoviVac from Russia).

Inevitably, along with the rise of different vaccines comes disinformation especially on social media. With the continuous rise in information emerging from online sources, the Internet, and social media specifically, have proven to be the main sources of vaccine-related information, as well as their mis- and dis-information. According to Himelein-Wachoviak et al., "As the virus spread across the United States, media coverage and information from online sources grew along with it. Among Americans, 72% report using an online news source for COVID-19 information in the last week, with 47% reporting that the source was social media." In fact, the World Health Organization estimates that as of April 2021, "nearly 6000 people around the globe were hospitalized because of coronavirus misinformation…[and] at least 800 people may have died due to misinformation related to COVID-19."

We currently live in the so-called "Information Age", and in such a time of mis- and dis-information, it is vital to understand the effects of the largest agent of such influences on the general public. Moreover, more urgency is placed on this study due to the ongoing pandemic. This being said, to examine the effects of COVID-19 information, including social media echo chambers, it is worth analyzing a set of science articles (albeit a relatively small number) and relating the sentiment polarity of these articles and determining whether or not a correlation is present between article sentiment score and COVID-19 vaccination rates in the US.

**Background**

It is important to note that another major player in curbing the spread of COVID-19 is the extent of vaccine hesitancy amongst the general public. Disinformation raises several issues on this, most prominently with regards to efforts on increasing vaccination rates. In fact, it is foreign disinformation which plays a more significant role in hindering vaccinations. From a report by Wilson and Wiyosonge, they concluded that foreign disinformation is correlated to a negative reception of COVID-19 vaccination information. "The substantive effect of foreign disinformation is to increase the number of negative vaccine tweets by 15% for the median country… there is a substantial relationship between foreign disinformation campaigns and declining vaccination coverage."

In another study published on June 25, 2022 by Rahmanti et. al, researchers showed that sentiments on Twitter were consistent with the sentiments of the general public with regards to COVID-19 vaccine safety. They showed that "there was a statistically significant trend of vaccination sentiment scores, which strongly correlated with the increase of vaccination coverage."

There are actually two methods for sentiment analysis — lexicon-based and machine learning-based — with their respective pros and cons. Lexicon-based sentiment analysis utilizes a dictionary or corpus of words which serves as the basis for text pre-processing. That being said, it also makes use of data/text pre-processing (removal of stopwords and punctuation). Here, researchers are mainly concerned with words either which denote positivity or negativity. It is worth noting that lexicons essentially map a certain emotive word with a particular value, often ranging between -1 to 1 (with -1 being very negative and +1 being very positive). It is also worth noting that superlatives can effectively "boost" or increase the resulting "sentiment score" or polarity of a body of text.

In a study by Reshi et al, researchers used various lexicon-based methods for sentiment analysis such as VADER and TextBlob. They showed that among the thousands of tweets gathered and processed, "TextBlob gives 12% negative tweets, while VADER assigns a negative polarity score to 22%..." Moreover, when VADER was used to conduct sentiment analysis the researchers found that compared to TextBlob, "there is a substantial change in the ratio of neutral and negative tweets. For example, the ratio of neutral tweets was changed from 48.81% to 37.74% for VADER, while negative tweets were raised to 22.31% from 12.86%."

Reshi et al. also used AFINN, another lexicon-based method of sentiment analysis and found out that from their dataset, the percentage of tweets which conveyed a negative sentiment differed depending on each lexicon-based approach (TextBlob, AFINN, and VADER) as shown in Figure 1 below.

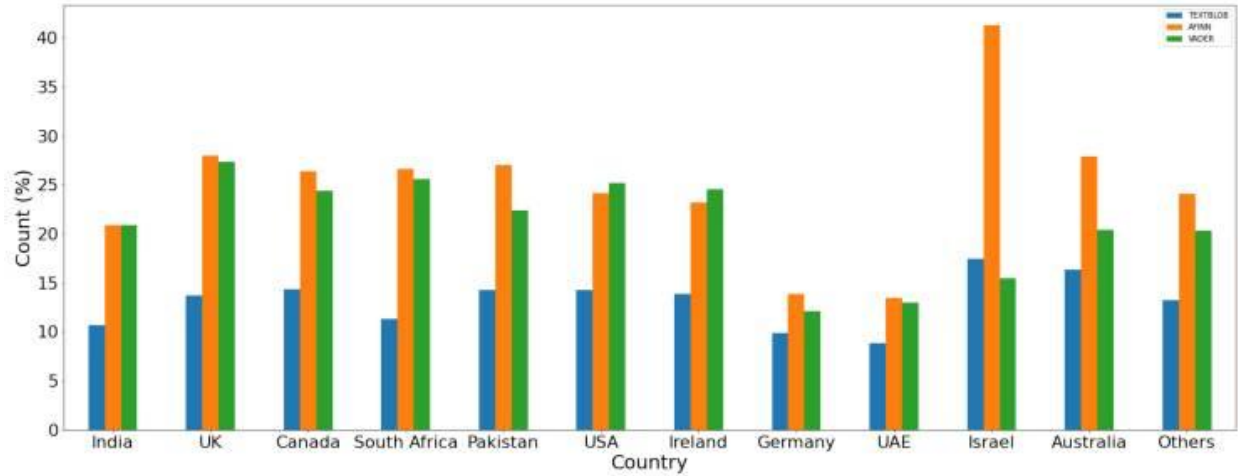

Figure 1: Negative Sentiment (%) Using Each Lexicon-Based Approach (Reshi et al., 2019)

Similar to lexicon-based sentiment analysis, machine learning-based Sentiment Analysis makes use of data/text pre-processing (removal of stopwords, punctuation, etc). However, instead of using a corpus or dictionary, ML-based sentiment analysis uses algorithms from deep learning and neural networks. A common neural network used is a convolutional neural network (CNN) which utilizes word embeddings — mappings of text onto multidimensional space which quantifies words based on their positivity or negativity (often ranging between -1 and 1). Each word is then placed into an embeddings matrix which serves as input to the CNN. Generally speaking, a number of Python libraries are also used to conduct sentiment analysis, albeit producing different results due to the different metrics that they consider (these include SpaCy and TextBlob (both of which are used for NLP, specifically in data/text preprocessing as shown in a study published by Pandey et. al. in 2019), as well as Tensorflow/Keras (more commonly used to create and customize training models, as well as easily use pre-existing common ML models, among many others).

In this research, we aim to investigate the effects of (albeit on a smaller scale) to determine whether or not general sentiments of published science articles from relatively reputable sources — "Health Day", "Popular Science", and "The Guardian" to name a few — have a large effect on the day-to-day increase or decrease in vaccination rates. The polarity/sentiment score of each article in the dataset was determined using VADER.

**Methodology - Data Preprocessing and Feature Engineering**

The main source of URLs processed or used in this analysis was taken from a paper entitled, "Identifying science in the news: An assessment of the precision and recall of Altmetric.com news mention data" (Fleecrackers et al., 2022). The main source of data on COVID-19 vaccinations in the US were taken from https://ourworldindata.org/covid-vaccinations based on an internationally-sourced dataset on GitHub, which can be found at https://github.com/owid/covid-19-data/tree/master/public/data.

In data pre-processing, the csv (comma-separated values) file was first obtained and "unnecessary" columns (columns which contained information not substantial or relevant to this study) were removed. These included the 'identifier', 'GeneralNotes', 'MentionNotes', 'gold_id'. NaNs found in the csv were also converted into blank strings for easier manipulation. The dates in the 'mention_date' column were converted into datetimes instead of strings in order to perform time-series forecasting later on as well.

In order to more efficiently obtain the text content of each article found in the aforementioned dataset, a scraper code was used. This scraper code used the BeautifulSoup library to parse through HTML source code (in this case, BeautifulSoup iterated through every <p> tag for each article to get text from every paragraph. A "for loop" was used to iterate through each article for text extraction. Unfortunately, some articles were put behind a paywall; hence incorrect or incomplete text in these articles was parsed. Afterward, a column called 'Content' within the original dataframe was created, and each article's parsed text was placed here.

Another piece of information that was extracted, aside from the text itself, was the "score" of each article. The total "score" of an article is how many times the following words appeared in the article: 'vaccine', 'vaccination', 'vaccinated', 'shot', 'shots', 'dose', 'doses', 'dosage', 'immunity', 'immune', 'immunization', 'immunizations'. Articles with a score of 0 were removed from the dataframe, as this had no overall impact to the dataset. Duplicate URLs were also removed so that each row contained a unique article URL).

**Methodology - Sentiment Analysis**

What is Sentiment Analysis? In a nutshell, it is a way to quantify how positive, negative, or neutral a certain text is. It gives a general idea of whether a positive or negative idea or feeling is being conveyed, despite a machine or computer not having an inherent idea of emotions or positivity/negativity.

In this study, VADER was used to obtain the necessary metrics used in analysis. VADER (Valence Aware Dictionary for Sentiment Reasoning) is an algorithm which concerns itself with polarities found in text. For example, the sentence, "The show was amazing" has a positive polarity while the sentence, "The show was terrible" has a negative polarity. As one can see, VADER can be used to determine how strong the positive or negative effect of a certain sentence or paragraph can be when it comes to COVID-19 vaccinations. VADER is a prominent example of lexicon-based sentiment analysis since it uses a corpus with existing values mapped to a large number of words. VADER is also considered lexicon-based also due to its consideration of stopwords, punctuation, and degree modifiers, all of which can alter the degree of positivity or negativity of a given text. Slang words are also considered in the computation of a text's sentiment score. It is worth mentioning that in ML-based sentiment analysis, text pre-processing is often performed in order to "strip down" a body of text to only its essential elements — which is suitable for a neural network but not for an algorithm like VADER. Some of VADER's advantages are that it can easily be applied, is efficient in dealing with a large amount of text such as those in articles, and it is relatively quick to implement, especially for the purposes of this research paper. However, sarcasm or irony, which can sometimes be found in articles, might skew or may cause inaccuracies in results due to misinterpretations.

In the context of this study, VADER was used to obtain the average sentiment score of all articles produced on a particular day. A 'Content' column was created within the main dataframe to extract the text from each article. VADER was then run through each of the cells in that column to obtain a compound score for the sentiment polarity of that article. Some articles had a "sentiment score" of 0, so they were removed as they had no impact on the consequent increase/decrease in daily vaccination rate margins. In this paper, the average sentiment score was obtained and used as one of the variables in analysis. This was determined by getting the mean of the sentiment scores of all articles published on a particular day.

**Results and Discussion**

Graph 1: Sentiment Score vs. Time

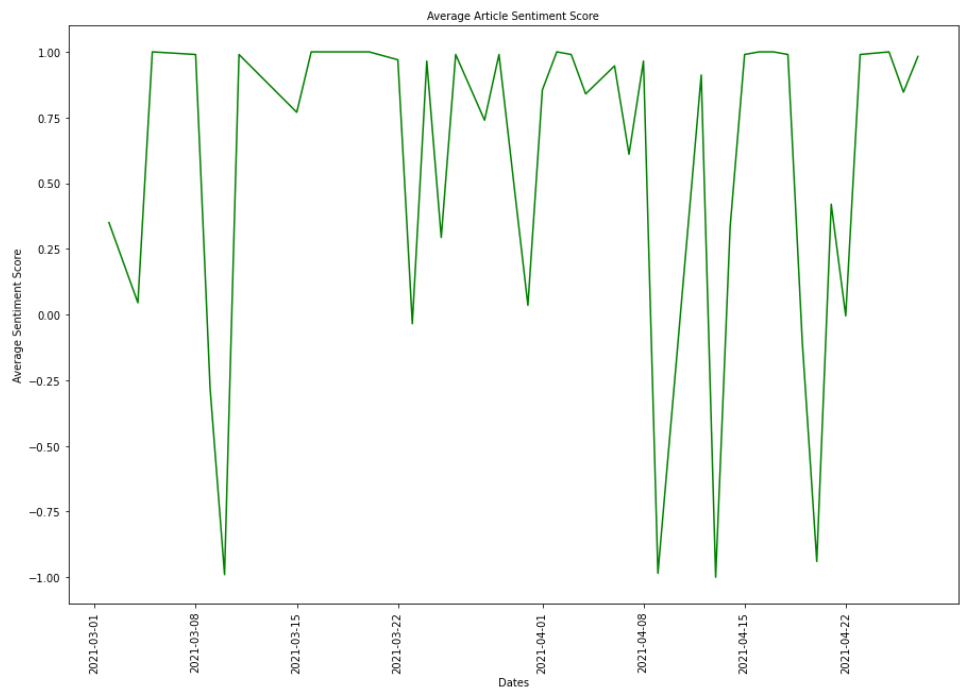

From Graph 1, it can be seen that on most days, the average sentiment score of published articles is ≥ 0; however, there are days wherein the average sentiment score reaches -1. Moreover, Graph 1 suggests the existence of periodicity, as there are noticeable decreases in the average sentiment score every few days.

The fluctuation of daily average sentiment scores may be due to specific events, or COVID-related breakthroughs or incidents which could have had either a positive or negative influence on the sentiment scores. Some examples of these events are headlined from various news sources below:

Date: 2021-03-05, Average Sentiment Score: 1.0
US Optimism About COVID-19 Situation Reaches New High; Backlash against Johnson & Johnson's COVID-19 vaccine is real and risky

Date: 2021-03-08, Average Sentiment Score: 0.99
The US Congress is preparing to send President Joe Biden an enormous $1.9 trillion economic stimulus and COVID-19 relief package

Date: 2021-03-15, Average Sentiment Score: 0.77
AstraZeneca COVID vaccine temporarily banned in growing list of countries - risk of blood clots; Covid-19 vaccine ads expected in next few weeks

Date: 2021-04-01, Average Sentiment Score: 0.855
Facebook is rolling out "I got my COVID-19 vaccine" profile frames; Human errors at a manufacturing plant forced Johnson & Johnson to throw out 15 million doses of its COVID-19 vaccine; Health Canada greenlights U.S. batch of AstraZeneca COVID-19 vaccine

Date: 2021-04-06, Average Sentiment Score: 0.9466666666666667
US insurers boost stakes in COVID-19 vaccine makers; U.S. job openings jump to two-year high in boost to labor; Pfizer/BioNTech, Moderna, and Johnson & Johnson may have more unexpected benefits to vaccination than experts initially saw coming.

Although this graph may mirror natural phenomena wherein vaccination rates experience a small dip every few days as well (as will be seen in Graph 3), the time period being analyzed is relatively short. To resolve this, a larger dataset with articles from a longer time period can be analyzed — analysis of weekly trends in vaccination rates could also be explored — and time series analysis using windows could be conducted (which will be discussed further in Graph 5).

Graph 2: Correlation of Sentiment Score vs. Time

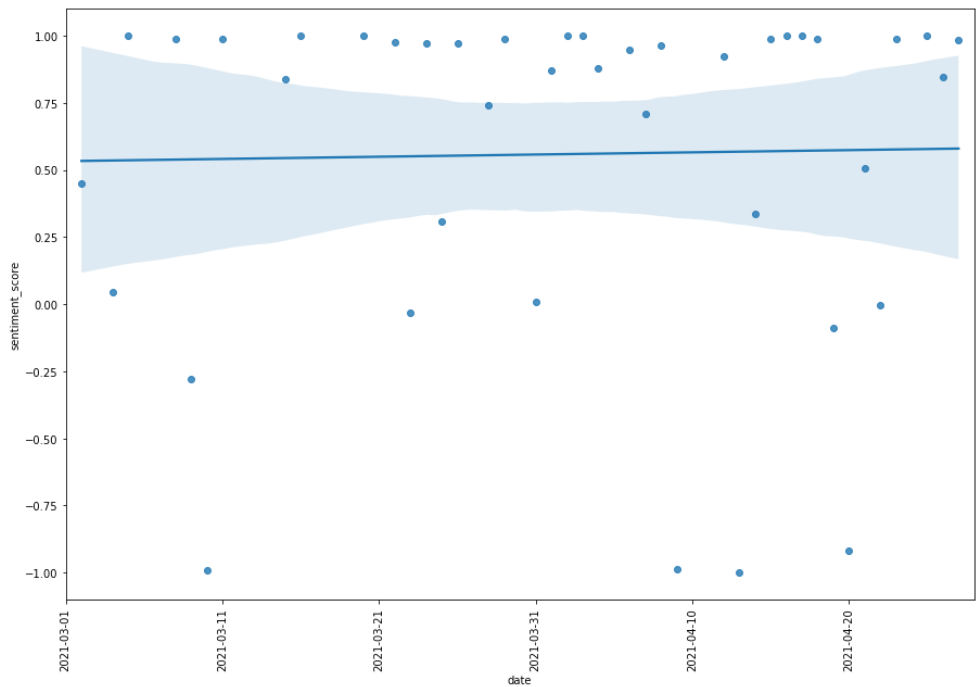

Here, the blue line represents the line of best fit based on all points in this graph. This suggests that the average sentiment scores of all articles within the dataset is slightly larger than 0.5 or that all articles generally carry a positive sentiment. That being said, as seen from the graph, the points generally have a weak correlation from each other; most of the data points are at a considerable distance from the line.

Graph 3: Increase in Total Percentage of Vaccinations vs. Time

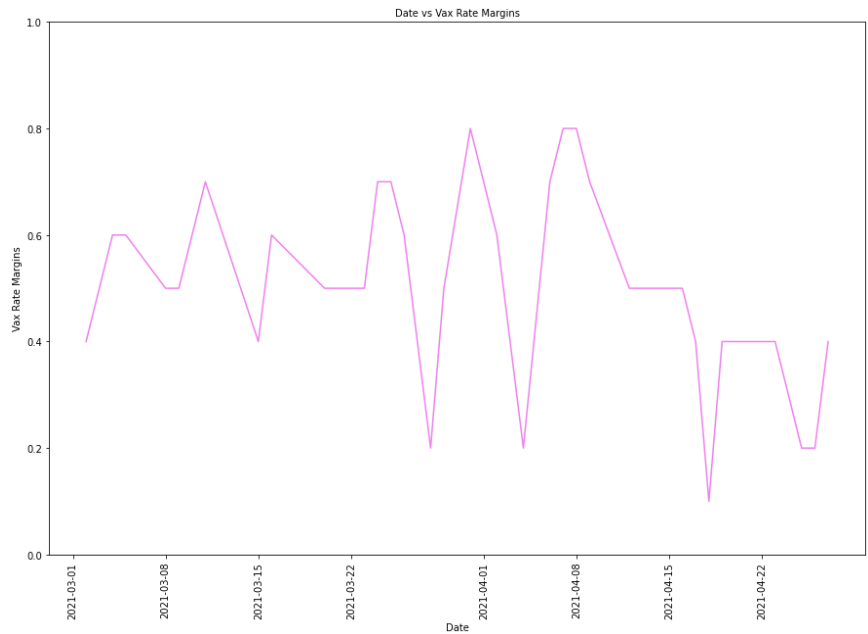

The dates of articles within the aforementioned dataset ranged from March 1, 2021 to April 27, 2021. It is worth noting that within this time period, the USA experienced an estimated additional increase of 3.6 million COVID-19 cases. The average sentiment score metric found in Graph 1 was obtained by calculating the mean of the sentiment scores of all articles published for a particular day. The day-to-day vaccination rate margins seen in Graph 2 were calculated by obtaining the difference between the cumulative percentage of people vaccinated for one day and that of the previous day.

Graph 4: Cross Correlation of Graphs 1 and 2

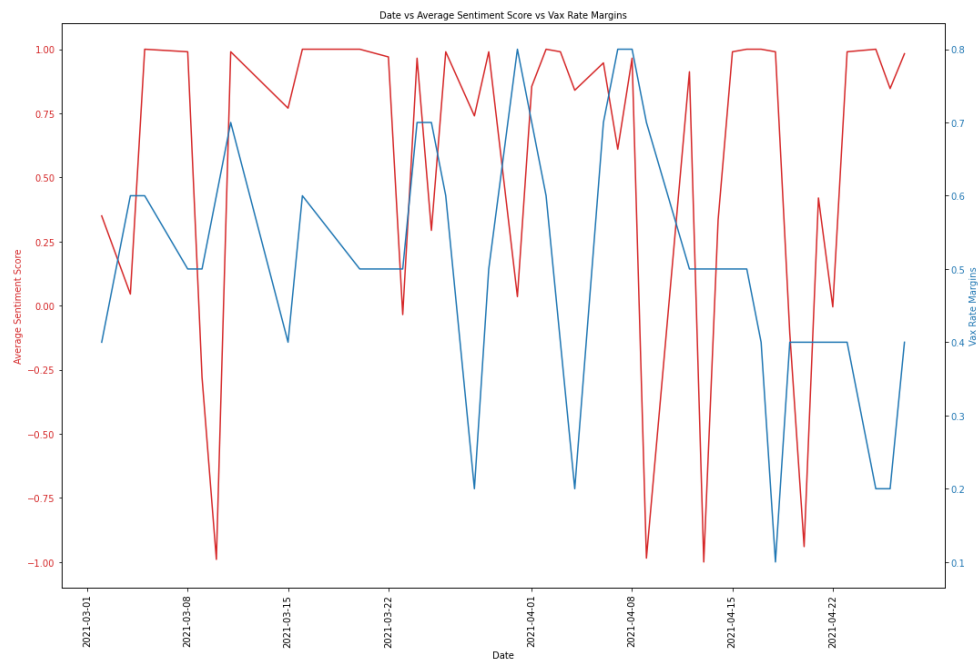

It is worth noting that the data obtained is not normally distributed. In addition, for the purposes of this study, Spearman's R Correlation serves as the basis for correlation instead of the Pearson Correlation. This is because the Pearson Correlation assumes homoscedasticity, which our data is not as data points are in differing distances from the line of best fit. Moreover, the Spearman Correlation coefficient can be used in non-linear associations between variables as seen in Graphs 1 and 2. With a Spearman's R correlation of around -8.5%, it can be said that there does not exist any significant correlation (or if there is, very weakly) between the average sentiment score of articles published on a particular day and the day-to-day percentage increase in vaccination rates.

It is worth noting, however, that the Spearman correlation coefficient is also a suitable metric to determine the monotonicity of the line of best fit. This was shown in a study done by Ye et al. (2015) where he and other researchers defined the Spearman correlation coefficient to be "a nonparametric (distribution free) rank statistic for evaluating the strength of monotone association between two independent variables." In their analysis of another study mentioned in the paper, they cited that some data (such as those involving two time-series) may be best analyzed — and the line of best fit measured — using the Spearman's correlation coefficient.

For comparison, the Pearson Correlation seen in Graph 3 is -0.12316. This suggests that when approximately by a linear "line of best-fit", there is a negative correlation of about 12% between the average sentiment score and the corresponding vaccination rate margins against time. The Spearman Correlation in Graph 3 is -0.08498. This suggests that the 2 metrics are very weakly monotonic; the line of best fit is very weakly decreasing.

Cross-correlation measures the relationship between 2 sets of time-series data or graphs, often with one graph being a "lagged" function of the second. The cross-correlation function (CCF) quantifies the correlation of a dataset with another dataset given a lag of *n*, where the value of *n* ranges from the desired beginning and ending time period being analyzed and/or predicted. In this study, the cross-correlation between the average article sentiment score and the corresponding increase or decrease in vaccination rate margins with lags from 0 to 40. This was done through the Python library statsmodels, which has already automated the calculation of the CCF. CCF values range from -1 to 1, or extremely negatively or positively correlated.

| Lag # | CCF | Lag # | CCF | Lag # | CCF | Lag # | CCF | Lag # | CCF |
|---|---|---|---|---|---|---|---|---|---|
| 0 | -0.1232 | 10 | -0.1409 | 20 | 0.0484 | 30 | -0.0632 | 40 | -0.4063 |
| 1 | 0.0289 | 11 | -0.2945 | 21 | -0.1006 | 31 | -0.0475 | | |
| 2 | 0.0991 | 12 | 0.0893 | 22 | -0.2463 | 32 | -0.1137 | | |
| 3 | -0.2030 | 13 | 0.2262 | 23 | -0.0758 | 33 | -0.0305 | | |
| 4 | 0.0972 | 14 | 0.1008 | 24 | -0.0508 | 34 | 0.2690 | | |
| 5 | -0.0555 | 15 | 0.0088 | 25 | 0.0827 | 35 | 0.0497 | | |
| 6 | 0.1686 | 16 | -0.0917 | 26 | 0.0142 | 36 | 0.2643 | | |
| 7 | 0.1166 | 17 | -0.1586 | 27 | 0.2528 | 37 | 0.0607 | | |
| 8 | -0.1037 | 18 | 0.2554 | 28 | -0.1575 | 38 | 0.0756 | | |
| 9 | -0.2238 | 19 | -0.0296 | 29 | -0.0117 | 39 | 0.0555 | | |

In the table above, it can be seen that in Lag 11, the CCF is weakly negatively correlated. This suggests that the daily average sentiment score and vaccination rate margin produce opposite reactions with each other. On the other hand, in Lags 18 and 34, the CCF is weakly positively correlated, which now suggests that the daily average sentiment score and vaccination rate margin produce similar reactions with each other.

**Conclusion**

Due to the rapid dissemination of information on the internet, most people get their information reading articles and social media posts. However, in the time of the COVID-19 pandemic, a number of scientific articles published by numerous reputable news websites have increasingly tackled COVID itself. In light of this, this study investigates and analyzes the effect of differing article sentiments on the corresponding increase or decrease in vaccination rates via the magnitude of daily vaccination rate margins. The analysis was performed using BeautifulSoup, which scrapes the text content from articles, and VADER, which evaluates the sentiment analysis given a certain text. Results show that there is only a weak correlation between article sentiment and vaccination rate margins.

**Limitations and Future Work**
The number of articles that ended up being processed was relatively small, which affected the analyses and findings gathered by this study. In addition, the articles processed stemmed from only a few sources which may have resulted in discrepancies between the actual and expected outcomes. All this being said, a deep learning-based approach instead of a lexicon-based one could be investigated further to better evaluate the accuracy of findings in this study and explore the different sentiments expressed by articles tackling different scientific concepts as well. In the study, about 2 months' worth of articles were processed; in light of this, the level of granularity for analysis could be changed to that over a longer time period — perhaps monthly or weekly —  to better understand and visualize long-term trends.

# Appendices

**Other Links**
- Complete list of articles used in dataset taken from Harvard Dataverse at https://dataverse.harvard.edu/dataset.xhtml?persistentId=doi:10.7910/DVN/WNDOFL
- Source Code for this paper can be found at https://colab.research.google.com/drive/1gJ8395TTlTfnLMtcH3Uw0s3HFyNewhAx?usp=sharing